\documentclass[10pt]{article}

\usepackage{graphicx,caption,subcaption}
\usepackage{pdflscape}
\graphicspath{{figures/}} 
\usepackage[labelsep=space]{caption}

\usepackage{booktabs, multirow} 
\usepackage{soul}
\usepackage{xcolor,colortbl} 
\usepackage{changepage,threeparttable} 

\usepackage{hyperref}

\usepackage{geometry}
\begin{document}

\title{\huge Comprehensive benchmarking of large language models\\ for RNA secondary structure prediction}
\author{L.I. Zablocki, L.A. Bugnon, M. Gerard, L. Di Persia, G. Stegmayer, D.H. Milone}

\date{Research Institute for Signals, Systems and Computational Intelligence\\ sinc(i), FICH-UNL, CONICET, Santa Fe, Argentina}

\maketitle

\begin{abstract}
    \noindent
    In recent years, inspired by the success of large language models (LLM) for DNA and proteins, several LLM for RNA have also been developed.
    These models take massive RNA datasets as inputs and learn, in a self-supervised way, how to represent each RNA base with a semantically rich numerical vector.
    This is done under the hypothesis that obtaining high-quality RNA representations can enhance data-costly downstream tasks, such as the fundamental RNA secondary structure prediction problem.
    However, existing RNA-LLM have not been evaluated for this task in a unified experimental setup. Since they are pre-trained models, assessment of their generalization capabilities on new structures is a crucial aspect. Nonetheless, this has been just partially addressed in literature.
    In this work we present a comprehensive experimental and comparative analysis of pre-trained RNA-LLM that have been recently proposed. We evaluate the use of these representations for the secondary structure prediction task with a common deep learning architecture. The RNA-LLM were assessed with increasing generalization difficulty on benchmark datasets. Results showed that two LLM clearly outperform the other models, and revealed significant challenges for generalization in low-homology scenarios. Moreover, in this study we provide curated benchmark datasets of increasing complexity and a unified experimental setup for this scientific endeavor.\\
    Availability: 
    The source code to reproduce all the experiments and results can be found in: \href{https://github.com/sinc-lab/rna-llm-folding}{https://github.com/sinc-lab/rna-llm-folding} \\
\end{abstract}

\section*{Introduction}

Ribonucleic acid (RNA) plays a crucial role in many fundamental biological processes, such as gene expression, cell signaling, and post-transcriptional regulation~\cite{Atkins2011, WaymentSteele2022}. As in proteins, RNAs function and interaction with other molecules are deeply related to their structure. For example, determining RNA structure is essential for RNA-based therapeutics such as mRNA vaccines~\cite{Pardi2018}. Among all RNA transcripts, only 5\% is responsible for protein coding. At the same time, a very large remaining portion is non-coding RNA (ncRNA)~\cite{Cech2014}, which in many cases adopt specific structures to perform important biological functions~\cite{Cao2024, amin2019}. Experimental results show that, to some degree, sequence determines the secondary structure, and thus, their function~\cite{Mortimer2014, Ganser2019, Vicens2022}. Therefore, the long-established and yet unresolved RNA secondary structure prediction is a major challenge in computational biology today~\cite{Bonnet2020,Bugnon2024}.

During many years, experimental molecular structure technologies like nuclear magnetic resonance, X-ray crystallography and cryogenic electron microscopy have produced several RNA structures~\cite{Hwang2024, Strobel2018}. However, despite the large number of ncRNA sequences available, most of their structures and functions remain still unknown~\cite{Yao2019}. At the same time that large amounts of unlabeled RNA sequence data were produced by high-throughput sequencing technologies, pre-trained RNA language models have started to be used for modeling the semantic space of RNA sequences, intending to facilitate its understanding. Motivated by the success of large language models (LLM) in proteins~\cite{min2021, rao2019, Heinzinger2019, asgari2015, alley2019, Rives2021, elnaggar2021, min2017} and DNA~\cite{liu2024}, several RNA LLM have recently appeared~\cite{Akiyama2022, rinalmo, Chen2022, Wang2023, Zhang2024, Yin2024, Wang2024} with the potential to be used for improving several important RNA-related tasks, among which one of the most relevant is RNA secondary structure prediction. 

RNA-LLM attempt to effectively embed RNA bases using deep representation learning, in particular the one developed in the field of natural language processing~\cite{Heinzinger2019,Rives2021}. Most RNA-LLM are based on bidirectional encoder representations from transformers (BERT), which was designed to generate context-sensitive distributed test token (word) representations~\cite{Devlin2019}. The relevance of LLM for transfer learning from sequences to downstream tasks have been analyzed in detail for proteins~\cite{unsal2022, Li2024, Vitale2024}.  In the case of RNA, the hypothesis is that LLM representation of nucleotide composition and sequence motifs can help characterize the structure and function of a sequence, analogously to how the meaning of a sentence is determined by the grammatical structure of natural language. Thus, word embedding techniques for natural language have been applied to bases for RNA sequences, obtaining a wide variety of BERT-based architectures trained on different databases. However, in spite that some theoretical reviews on the applications and utility of LLM in bioinformatics have recently appeared~\cite{liu2024, Szaata2024, Peng2024}, to the best of our knowledge all the available RNA-LLM have not yet been comparatively evaluated for the secondary structure prediction task, considering a fair experimental setup, with the same datasets and structure prediction model. Since LLM are pre-trained models, it is important to analyze how they were trained to assess generalization capabilities on new sequences/structures, considering homology-aware partitions and cross-family predictions.

\begin{table*}[t!]
    \caption{\textbf{RNA large language models included in this study.}}
    \newcommand{\colA}[1]{#1}
\newcommand{\colAb}[1]{\textbf{#1}}
\newcommand{\colB}[1]{#1}

\newcommand{\github}[1]{\href{#1}{\vspace*{-0.3mm}\includegraphics[height=2.5ex, trim=50mm 30mm 50mm 10mm, clip]{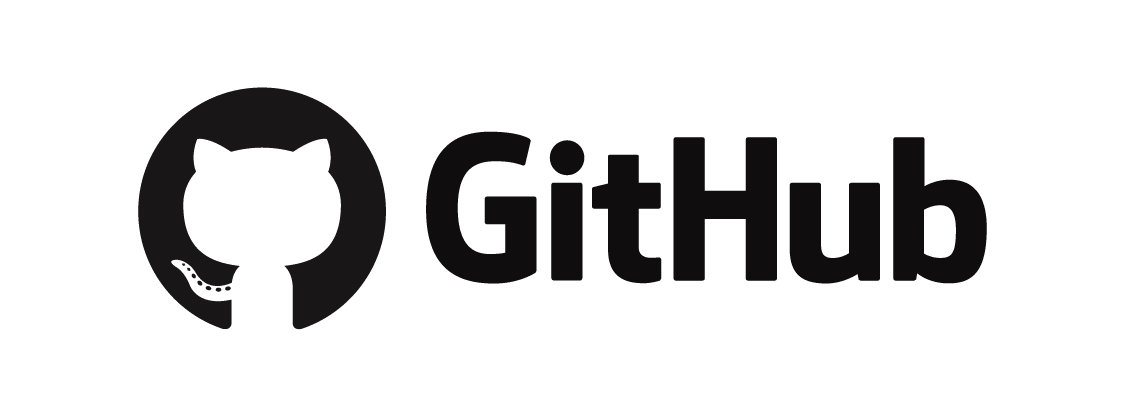}}}
\newcommand{\zenodo}[1]{\href{#1}{\vspace*{-0.3mm}\includegraphics[height=2.5ex, trim=50mm 30mm 50mm 10mm, clip]{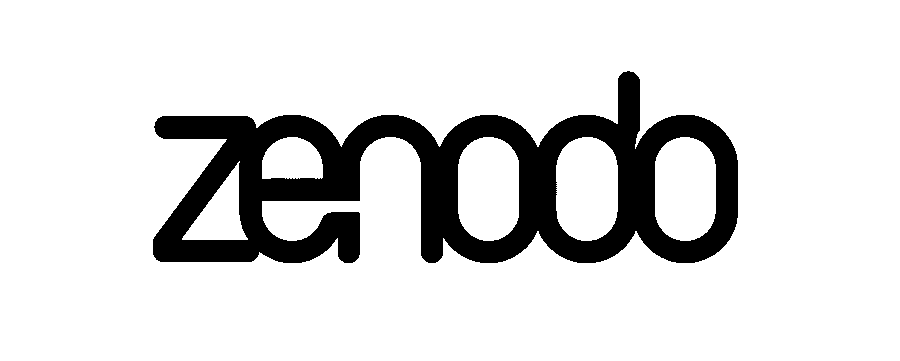}}}

\fontsize{6}{6}\selectfont

\begin{tabular}{lcrrccrc}
    \toprule
    \vspace*{-2.80mm} \\
    \colAb{RNA-LLM} & 
        \colAb{Year} &
        \colAb{Emb. Dim.} &
        \colAb{Pre-train seqs.} &
        \colAb{Databases} &
        \colAb{Architecture (layers)} &
        \colAb{Parameters} &
        \colAb{Repository} \\

    \vspace*{-2.61mm} \\
    \cmidrule{1-8}
    \vspace*{-3.00mm} \\
    RNABERT~\cite{Akiyama2022} &
        2022         &
        120 &           
        76,237 & 
        RNAcentral         &  
        Transformer (6)      &            
        509,896 & 
        \github{https://github.com/mana438/RNABERT} \\
    {}& \multicolumn{7}{l}{
        \begin{minipage}{150mm}
            \hrule
            \vspace*{0.5mm}
            Pre-training with masked language modeling (MLM)\cite{devlin} masking 15\% random bases and predicting from neighboring bases, with data augmentation taking 10 copies of each ncRNA and applying 10 different mask patterns.

            Structural alignment learning (SAL) pre-training task based on seed alignment for each RNA family with Rfam.

            Token embedding randomly generated 120 dimensions RNA bases.

            Positional embedding for each base in the sequence.

            Stack of transformers with multi-head self-attention and feed-forward neural network. 

            Softmax output layer.

            Cross-entropy loss function.
        \end{minipage}
   } \\
    \cmidrule{1-8}
    \vspace*{-3.00mm} \\
    \colB{
    RNA-FM~\cite{Chen2022}} &
        \colB{2022} & 
        \colB{640} &
        \colB{23,700,000} &
        \colB{RNAcentral} & 
        \colB{Transformer (12)} & 
        \colB{$\sim$100,000,000} & 
        \colB{\github{https://github.com/ml4bio/RNA-FM}}
        \\
    \colB{ 
    {}&\multicolumn{7}{l}{\hspace*{-0.8mm} 
        \begin{minipage}{150mm}
           \hrule
           \vspace*{0.5mm}
           Pre-training with MLM and 15\% random masking.
           
           Based on BERT original architecture.
           
           Transformer encoder blocks, with multi-head self-attention modules and feed-forward layers.
           
           Softmax output layer.
           
           Cross-entropy loss function.						
        \end{minipage}
    }} \\
    \vspace*{-2.61mm} \\
    \cmidrule{1-8}
    \vspace*{-3.00mm} \\
    RNA-MSM~\cite{Zhang2024} &
        2024         &          
        768 &           
        3,087,138 & 
        Rfam         &  
        Transformer (12)      &            
        $\sim$96,000,000 & 
        \github{https://github.com/yikunpku/RNA-MSM} \\
    {}& \multicolumn{7}{l}{
    \begin{minipage}{150mm}
        \hrule
        \vspace*{0.5mm}
        Multiple sequence alignment (MSA) inspired by AlphaFold2 \cite{Jumper2021}.

        Pre-training with MLM and 20\% random masking.
        
        Homology search and MSA of training sequences with RNAcmap3 database \cite{Chen2024} as augmented input.
        
        Two modules: embedding and MSA transformer.
        
        Initial embedding layer and two learnable position-embedding layers.
        
        Stack of MSA transformer blocks with a residue and sequence attention layer followed by a feed-forward layer with GELU activation functions.
        
        Cross-entropy loss function.
    \end{minipage}
    } \\
    \cmidrule{1-8}
    \vspace*{-3.00mm} \\
    \colB{
    ERNIE-RNA~\cite{Yin2024}} &
        \colB{2024} & 
        \colB{768} &
        \colB{20,400,000} &
        \colB{RNAcentral} & 
        \colB{Transformer (12)} & 
        \colB{86,000,000} & 
        \colB{\github{https://github.com/Bruce-ywj/ERNIE-RNA}} \\
    \colB{ 
    {}&\multicolumn{7}{l}{\hspace*{-0.8mm}
        \begin{minipage}{150mm}
           \hrule
           \vspace*{0.5mm}
           Pre-training with MLM and 15\% random masking.
           
           Based on the Enhanced Representation through Knowledge Integration (ERNIE) framework~\cite{Sun2019}.
           
           Modified BERT that incorporates base-pairing informed attention bias when calculating attention.
           
           From the second layer onward, the bias of each layer is determined by the attention map of the previous layer.
           
           Transformer blocks with attention heads followed by a feed-forward layer.
           
           Cross-entropy loss function.
        \end{minipage}
    }} \\
    \vspace*{-2.61mm} \\
    \cmidrule{1-8}
    \vspace*{-3.00mm} \\
    RNAErnie~\cite{Wang2024} &
        2024         &          
        768 &           $
        \sim$23,000,000 & 
        RNAcentral         &  
        Transformer (12)      &            $
        \sim$105,000,000 & 
        \zenodo{https://zenodo.org/records/10847621} \\
    {}& \multicolumn{7}{l}{
        \begin{minipage}{150mm}
            \hrule
            \vspace*{0.5mm}
            Pre-training with MLM and 15\% random masking.
    
            Motif-aware additional pre-training strategy involving motif-level random masking from motif databases \cite{Leontis2006}, and random subsequence masking.
    
            Built upon the ERNIE framework.
    
            Transformer blocks followed by a feed-forward layer.
    
            Cross-entropy loss function.
        \end{minipage}
    } \\
    \cmidrule{1-8}
    \vspace*{-3.00mm} \\
    \colB{
    RiNALMo~\cite{rinalmo}} &
        \colB{2024} & 
        \colB{1280} &
        \colB{$\sim$36,000,000} &
        \colB{RNAcentral + Rfam + Ensembl} & 
        \colB{Transformer (33)} & 
        \colB{$\sim$650,000,000} & 
        \colB{\github{https://github.com/lbcb-sci/RiNALMo}} \\
    %
    \colB{ 
    {}&\multicolumn{7}{l}{\hspace*{-0.8mm}
        \begin{minipage}{150mm}
        \hrule
        \vspace*{0.5mm}
        Pre-training with MLM and 15\% random masking and FlashAttention-2 \cite{dao2023}.
        
        BERT-style encoder-only transformer.
        
        Position encoded using rotary positional embedding \cite{su2024}.
        
        Each block comprises a multi-head attention of 20 heads.
        
        Feed-forward network with two linear layers and SwiGLU activation functions \cite{shazeer2020}.
        
        Residual connections with layer normalization to stabilize training.
        
        Cross-entropy loss function.
        \end{minipage}
    }} \\
\vspace*{-2.00mm} \\
\bottomrule
\end{tabular}
    \label{tab:llms}
\end{table*} 

In this work, we provide a comprehensive experimental comparison of the latest pretrained RNA-LLM, using a unified and consistent experimental setup. It offers an independent, third-party assessment following the best practice guidelines for bias-free evaluations in the community~\cite{Zhang2022}. We provide a description of each model, together with an experimental testing in four benchmarks of increasing complexity, with the corresponding source code and datasets to ensure reproducibility, supporting the development and boosting of future improvements on RNA secondary structure prediction based on LLM.

%
%
\section*{RNA large language models evaluated}
In the last three years, a number of RNA-LLM have appeared in literature.
The following RNA-LLM (summarized in Table~\ref{tab:llms}) were included in this study according to their availability as open source tools.

\textbf{RNABERT}~\cite{Akiyama2022}: For the pre-training of the masked language modeling (MLM)\cite{devlin} task in this LLM, $76,237$ human derived small ncRNAs from RNAcentral~\cite{rnacentral} were utilized. First, a token embedding randomly generates a 120-dimensional numerical vector that encodes the four RNA bases (A, C, G, U) and assigns the same vector to each base in the input RNA sequence. Second, the position embedding generates a 120-dimensional vector that encodes the position information of each base in the sequence. Third, the element-wise sum of token embedding and position embedding for each base in the input RNA sequence is fed to the Transformer layer. RNABERT architecture consists of a stack of 6 Transformers, each of which is composed of a multi-head self-attention mechanism followed by a feed-forward neural network. The weights of the last layer are trained by alternating between different tasks. The MLM task masks 15\% of the bases randomly selected from the input RNA sequence and predicts the masked part using the neighboring bases, enabling a context sensitive embedding. The structural alignment learning (SAL) task is based on RNA structural alignment and aims to obtain closer embeddings for bases in the same column of reference alignment. The seed alignment for each family was downloaded from Rfam~\cite{Kalvari2017} as the reference structural alignment. The MLM task enables the position and context-sensitive embedding, and the SAL task enables the structural information embedding.  

\textbf{RNA-FM}~\cite{Chen2022}: It is a 100 million parameter foundation model encoder based on the BERT original implementation and trained on 23.7 million unannotated ncRNAs from RNAcentral~\cite{rnacentral} database. Identical sequences were removed by applying CD-HIT~\cite{Fu2012} with a cut-off at 100\%. The resulting dataset was used to train the foundation model in a self-supervised manner, reconstructing the input masked tokens as a pretext task. During pre-training under self-supervised training, around 15\% of nucleotide tokens were randomly replaced with a special mask token. RNA-FM takes raw sequential tokens as input and embeds each nucleotide into a 640-dimensional vector. The architecture has 12 transformer encoder blocks as in BERT, which includes multi-head self-attention modules and feed-forward layers, with a final softmax layer to predict the output.  The model was trained with MLM by predicting the original masked token with cross-entropy loss.  

\textbf{RNA-MSM}~\cite{Zhang2024}: Inspired by the success of AlphaFold2~\cite{Jumper2021} and the use of homologous sequences for the highly accurate prediction of protein structures, RNA-MSM is a multiple sequence alignment (MSA)-based RNA language model. It utilizes a set of homologous sequences that allows having a larger number of sequences for training. $4,069$ RNA families were downloaded from Rfam 14.7, totaling $3,087,138$ sequences. The RNAcmap3 database~\cite{Chen2024} for homolog search and sequence alignment was employed. Rfam families containing RNA sequences with experimentally determined structures were excluded to minimize potential overfitting for downstream tasks such as structural inference. This led to a total of $3,932$ Rfam families. 
Pre-training was based on MLM, with 20\% random masking.
The embedding module, with one initial embedding layer and two learnable position-embedding layers, encodes entries in the MSA separately. The architecture is made of a stack of MSA transformer blocks, each with a residue and sequence attention layer containing 12 heads with an embedding size of $768$, followed by a feed-forward layer, which summarizes approximately 96 million parameters. The use of structural information associated with the sequences was not included during the LLM pre-training. 
To provide a fair comparison with the other RNA-LLM, a single-sequence version of this method was used. This way, the embedding obtained corresponds only to the input sequence.

\textbf{ERNIE-RNA}~\cite{Yin2024}: This RNA pre-trained language model is based on the Enhanced Representation through Knowledge Integration (ERNIE) framework~\cite{Sun2019} and a modified BERT that incorporates base-pairing restrictions to be used with RNA\@. For training the model, a 34 million ncRNA dataset from the RNAcentral database was downloaded. After refining the vocabulary and removing redundant sequences, the final dataset consisted of 20.4 million RNA sequences. ERNIE-RNA was trained with MLM, which predicts the masked token with cross-entropy loss. The architecture has 12 transformer blocks, with 12 attention heads each. Every token in the input sequences is mapped to a $768$-dimensional vector, resulting in 86 million parameters.  The main hypothesis is that ERNIE-RNA can learn functional and structural information thanks to the use of attention maps during pre-training.

\textbf{RNAErnie}~\cite{Wang2024}: It is also built upon the Enhanced Representation through Knowledge Integration (ERNIE) framework~\cite{Sun2019}, together with multilayer and multihead transformer blocks. Pre-training was done with a corpus of approximately 23 million sequences extracted from the RNAcentral, and using self-supervised learning with multilevel random masking. The main difference with other works was the use of a motif-aware pre-training strategy involving motif-level and subsequence random masking, which can capture both subsequence and motif-level knowledge extracted from motif databases~\cite{Leontis2006}. The architecture of RNAErnie shares the same architectural configuration as ERNIE 2.0~\cite{Sun2020}: a 12-layer transformer with a hidden state embedding dimension of $768$. A block first tokenizes RNA bases in the sequence and subsequently feeds 	them into the transformer. Given the embeddings for every token in the RNA sequence, the RNAErnie basic block transforms the series of token embeddings into a $768 \times L$ embedding using trainable parameters and then outputs the embedding of the RNA sequence. The total number of trainable parameters in RNAErnie is approximately 105 million.

\textbf{RiNALMo}~\cite{rinalmo}: It is the largest RNA language model to date with 650 million parameters pre-trained on 36 million unique non-coding RNA sequences from the RNAcentral~\cite{rnacentral} database augmented by Rfam~\cite{Kalvari2021}, nt~\cite{Sayers2022} and Ensembl~\cite{Martin2022}. To ensure diversity in each batch, the sequences were clustered with MMSeqs2~\cite{Steinegger2017} into 17 million clusters, and then each batch contained a mixture of sequences sampled from different clusters. The architecture of RiNALMo is a BERT-style encoder-only Transformer~\cite{devlin}. Before passing to the Transformer, an RNA sequence is tokenized and represented as a $1280$-dimensional vector. RiNALMo consists of 33 Transformer blocks, where each block comprises a multi-head attention and a feed-forward network. The position of the tokens is encoded using rotary positional embedding~\cite{su2024}. Each multi-head attention has 20 heads. To improve pre-training efficiency, FlashAttention-2~\cite{dao2023} is employed. In the feed-forward network, two linear layers are used together with SwiGLU activation function~\cite{shazeer2020}. Among the transformer modules, there are residual connections with layer normalization to stabilize training. 

\section*{Materials and methods}
\subsection*{Data}
The following datasets have been used in this study.
For all datasets, sequences longer than $512$ nucleotides were filtered to limit computational requirements~\cite{Szikszai2022}. That is, $512$ nt is the top length in each dataset.

\textbf{ArchiveII dataset}~\cite{Sloma2016}: The most widely used benchmark dataset for RNA folding methods, containing RNA structures from 9 RNA families: 5s (ribosomal RNAs), srp (signal recognition particle), tRNA (transfer RNA), tmRNA (transfer messenger RNA), RNaseP (Ribonuclease P), grp1 (Glycine-rich RNA-binding protein 1), 16s (ribosomal RNA), telomerase and 23s (ribosomal RNA). The total number of sequences is $3,864$.  This dataset is used in two training and testing split configurations. First, a random set of 5-fold partitions, following the original splits provided by the authors for the ArchiveII dataset~\cite{Sloma2016}. We also perform a cross-family generalization analysis, training on all RNA families but one that is used as a test unseen family, and repeating for all families. That is, a leave-one-family-out strategy.

\textbf{bpRNA dataset}~\cite{Singh2019}: The same train and test sets as used in SPOT-RNA\cite{Singh2019}. It is a non-redundant set of RNA sequences at 80\% sequence-identity cutoff with CD-HIT-EST, with annotated secondary structure from bpRNA34~\cite{Fu2012}. This filtered dataset of $13,419$ RNAs is randomly divided into $10,814$ RNAs for training (TR0), $1,300$ for validation (VL0), and $1,305$ for an independent test (TS0). In this dataset, each model was trained with TR0+VL0 and tested with TS0. 

\textbf{bpRNA-new dataset}~\cite{Singh2019}: This dataset was derived from Rfam 14.2~\cite{Kalvari2021}, containing new RNA families different from the bpRNA dataset. This test dataset has $5,401$ sequences and was built to assess cross-family model generalization. This dataset is used as an additional test set for the models trained with the bpRNA dataset.

\textbf{PDB-RNA dataset}
~\cite{Sato2021}
The PDB sets TR1 (training), VL1 (validation) and TS1 (testing) are the same sets as in SPOT-RNA, prepared by downloading all the high-resolution ($<3.5$A\textdegree) RNA X-ray structures from the PDB dataset on March 2, 2019. The numbers of structures for TR1, VL1 and TS1 are $120$, $30$ and $62$, respectively, after removing homologous sequences between and within the sets by CD-HIT-EST at the lowest allowed sequence identity cut-off of 80\%. In this dataset each model was trained with TR1+VL1 and tested with TS1. This small training set is used to test the LLM capabilities in a very hard setup for any transfer learning approach.

\subsection*{Self-supervised RNA-LLM and prediction model}

\begin{figure*}[p]
    \centering
    \includegraphics[width=177mm]{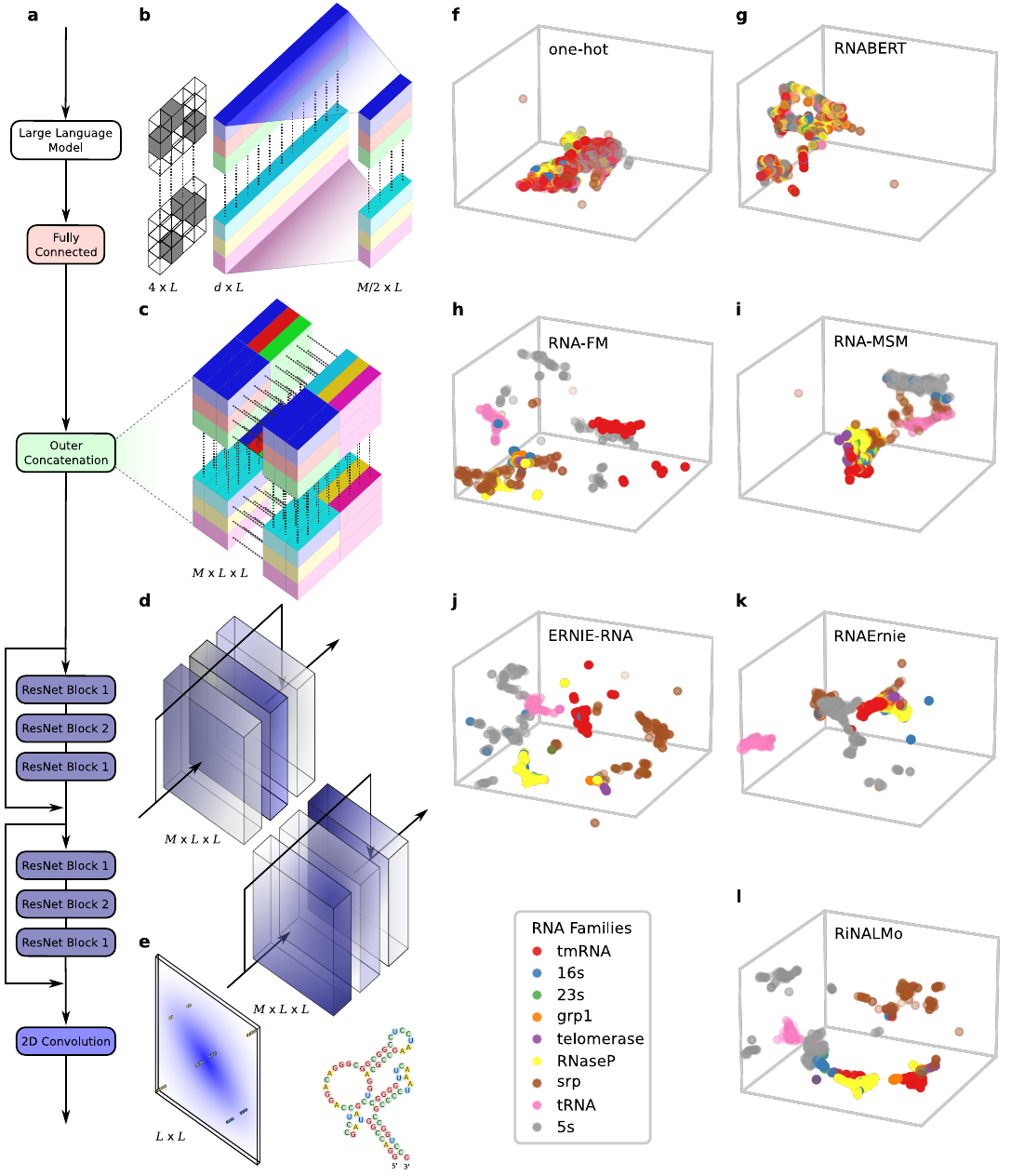}
    \caption{RNA-LLM embeddings and deep neural architecture of the prediction model.
    \textbf{a}, Flow diagram for RNA secondary structure predictions. Each LLM was downloaded from its official repository, frozen and used to get per-nucleotide embeddings for each sequence in the benchmark datasets. The embeddings go through a fully connected layer, outer concatenation, two 2D ResNet blocks and a final 2D convolution. This flow is explained in detail in aligned panels b-e. 
    \textbf{b}, One-hot encoding, of size $4 \times L$, and LLM representation  with size $d \times L$, where $d$ is the embedding dimension and $L$ is the padded sequence length. A fully connected layer reduces the input dimension to $M/2$. 
    \textbf{c}, The $M/2 \times L$ projection is transformed to a $M \times L \times L$ tensor using outer concatenation. The position $(i, j)$ contains the concatenated representation of nucleotides $i$ and $j$.
    \textbf{d}, Then, there are two 2D ResNet blocks with 2D Convolution, Instance Normalization and ReLU\@. ResNet Blocks 1 has kernel size 1 and ResNet Blocks 2 has kernel size 3.
    \textbf{e}, A final convolution yields the output scores as a $L \times L$ connection matrix.
    \textbf{f-l}, RNA-LLM embeddings projected with UMAP for dimensionality reduction. The RNA families of the ArchiveII dataset are highlighted with different colors. Sequences within the same RNA family are expected to be close in the dimensionally reduced space.}%
    \label{fig:predictor}
\end{figure*}

For the comparative analysis, embeddings of each RNA-LLM feed the same deep learning (DL) architecture (Figure~\ref{fig:predictor}a-e) for RNA secondary structure prediction. 
The rationale behind this is that it should not be necessary to optimize the classifier hyperparameters for each LLM, since those have complex architectures and have already been trained with a large amount of data. Thus, according to the transfer learning paradigm, the embeddings should have enough information to solve the downstream task. Moreover, any difference measured in performance will be due to the LLM and not because of the classifier architecture. This is the only trainable part of the experimental setup, which uses the train-test partitions defined for each benchmark dataset. 

Sequence one-hot embedding was defined as usual using 4 positions for A, U, G and C (Fig.~\ref{fig:predictor}b). Each LLM was used with the pre-trained weights provided, according to instructions in the official repositories. Models were frozen, that is, not re-trained nor fine-tuned. Per-nucleotide embeddings were extracted for each sequence in each benchmark dataset, obtaining a $d \times L$ tensor, with $d$ the embedding dimension and $L$ the sequence length. The embeddings feed the secondary structure prediction network described in Figure~\ref{fig:predictor}a. This architecture was designed following the one used in RiNALMo~\cite{rinalmo}, RNA-FM~\cite{Chen2022}, RNA-MSM~\cite{Zhang2024}, and ERNIE-RNA~\cite{Yin2024}. A fully connected layer reduces the input dimension to $M/2$ (being $M \ll d$), in order to obtain the same dimension for all RNA-LLM\@. An outer concatenation approach was used to transform the $M/2 \times L$ projection to a $M \times L \times L$. The $M$ dimension can be interpreted as channels of a $L \times L$ image. Then, this tensor passes through two 2D ResNet blocks~\cite{He2016} with the following configuration: a first ResNet with convolutions of kernel size 1 and a second one with kernel size 3. Both ResNet blocks include instance Normalization and ReLU activation function. The same DL architecture was trained and tested using each RNA-LLM and dataset. For training, Adam optimizer was used with a learning rate of $0.0001$, a batch size of 4 and binary cross-entropy as the loss function. A fixed number of 15 epochs was used in order that all RNA-LLM have equal possibilities for training. The number of epochs and the learning rate were determined in preliminary experiments, ensuring that there was no overfitting nor underfitting of the secondary structure prediction model for any of the RNA-LLM embeddings used as input. Each dataset was used with its own training and testing partitions, except for bpRNA-new, which was only used as a testing partition for the prediction model already trained on bpRNA\@. All experiments were run on 3 servers, each with 2 NVIDIA RTX A5000 GPU.

For the comparison of  RNA-LLM based predictors with baseline methods, the most widely-used pure-thermodynamic structure predictors were included: RNAstructure \cite{Reuter2010}, RNAfold \cite{Lorenz2011} and LinearPartition-V \cite{Zhang2020}. We have also incorporated hybrid methods, that is, a mix of thermodynamic and machine learning approaches, such as LinearPartition-C \cite{Zhang2020} and MXFold2 \cite{Sato2021}. Finally, we also included the most recent pure DL prediction methods: UFold \cite{Fu2022},  REDfold \cite{Chen2023}, and sincFold \cite{Bugnon2024}.

\subsection*{Performance measures}
Results are reported with the following performance measures.

\textbf{Base pairs metrics:} The focus of performance measures is on the predicted base pairs in comparison to a reference structure~\cite{Mathews2019}. Pairs that are both in the prediction and the reference structure are true positives (TP), while pairs predicted but not in the true structure are false positives (FP). Similarly, a pair in the reference structure that is not predicted is a false negative (FN), and a pair that is neither predicted nor in the true structure is a true negative (TN). 
A widely used metric is the recall (or sensitivity), defined as the ratio of TP to all the true pairs (TP+FN). It is a measure of how many predicted pairs are true. 
Insted, the precision is defined as the ratio of TP to all the predicted pairs (TP+FP). This relation between TP and FP is very important in the context of class imbalance, because FP can be a large number in comparison to TP, and this is not reflected in the recall.
The $F_1$ score is the harmonic mean of precision and recall, thus summarizing both measures of performance in a single value. Therefore, in this work the $F_1$ score is used as the global performance measure for the comparison of the different methods. It is defined as $F_{1}=\frac{2 TP}{2 TP + FP + FN}$.

\textbf{Structural motifs:} Stems, multiloops, internal loops, bulges, hairpin loops, dangling ends, and external loops were extracted using the bpRNA toolkit~\cite{Danaee2018} on the reference dot-bracket structures. All reference and predicted base-pairs were extracted for positions of each motif type, and average $F_1$ computed separately. 

\textbf{Structural similarity:} to complement the $F_1$ score as an evaluation metric, we have also calculated a structural measure, the Weisfeiler-Lehman graph kernel (WL) metric~\cite{runge2023}. The WL metric first assigns to each node (nucleotide) in the graph (secondary structure) a label representing its local structural information. Then, a label propagation step iterates over the nodes and updates their labels based on the labels of their neighboring nodes. Finally, a hash function is computed that aggregates these labels to generate a feature vector. The WL is defined as inner product $WL(G_1,G_2) = \Phi(G_1) \cdot \Phi(G_2)$, where $\Phi(G_i)$ represents the feature vector of graph $G_i$ obtained by aggregating the labels through the hash function.

\section*{Results}

\subsection*{LLM separate structural families even when they were not pre-trained with family information}

A useful RNA embedding for the RNA secondary structure prediction task is expected to well-represent not only sequence properties, but also structural aspects. Therefore, RNA embeddings encoding similar features should share a common region in the multidimensional embedding space. 
To have a preliminary insight into this topic, a qualitative comparison of LLM embeddings of the ArchiveII dataset was performed by nonlinearly reducing them to three dimensions using the uniform manifold approximation and projection (UMAP)~\cite{McInnes2018}. After that, each RNA sequence was depicted with a different color according 
to its corresponding RNA family (Fig.~\ref{fig:predictor}f-l).

The one-hot encoding (Fig.~\ref{fig:predictor}f) was used as reference. In this case, as for RNABERT (Fig.~\ref{fig:predictor}g) representations, there is a large overlap and mixing of sequences that belong to different RNA families; that is, in the same close region there are sequences that are very different in length, structure, and function. Instead, in the case of RNA-FM (Fig.~\ref{fig:predictor}h), ERNIE-RNA (Fig.~\ref{fig:predictor}j) and RiNALMo (Fig.~\ref{fig:predictor}l) each RNA family is mostly separated from the others. The 5s family (gray), which is the largest family, is distributed in several regions of the projected space. In the case of RNA-FM, the 5s, tmRNA (red) and tRNA (pink) families are well-separated among them. Overall, it can be stated that for RiNALMo and ERNIE-RNA, most RNA sequences and families are well-separated in the 3D projected space. In the case of RNA-MSM (Fig.~\ref{fig:predictor}i), there are two large and separated clusters but with a mix of families within. One large cluster has tmRNA, RNaseP (yellow), srp (brown), 23s (green), and telomerase (violet); while the other cluster has 5s, srp and tRNA\@. RNAErnie has a mix of very well-separated and cohesive groups of sequences belonging to the families tRNA, srp and 5s; but there is also a large overlap among sequences of the other families. The telomerase family, the longest one, can hardly be distinguished in any case, being cohesively and separately represented only in the case of RiNALMo and ERNIE-RNA\@.

\subsection*{Performance on homology-challenging datasets}

\begin{figure*}[t!]
    \centering
    \includegraphics[width=176mm]{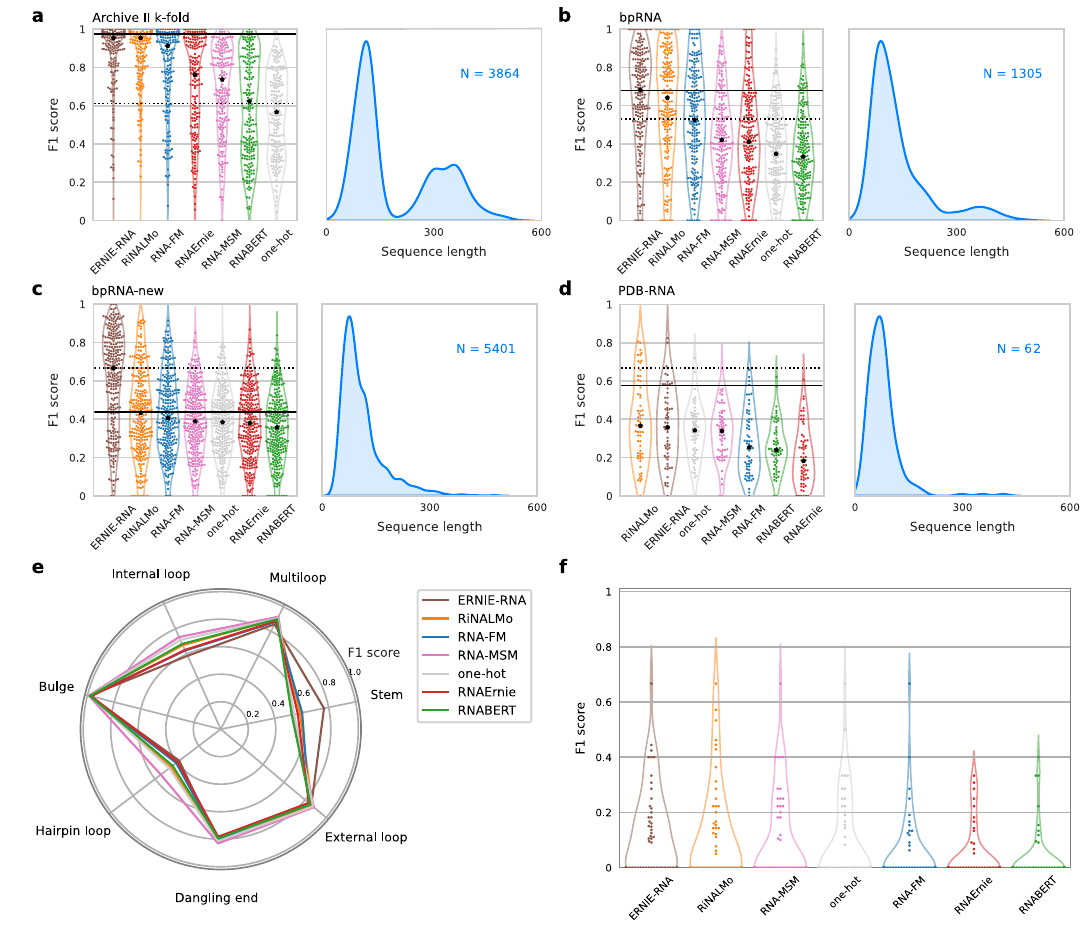}
    \caption{Comparative results among RNA-LLM on the RNA secondary structure prediction task for different benchmark datasets of increasing complexity. 
    Each method has a different color.
    The thermodynamic prediction method (LinearPartition-V, dashed black) and the DL-based prediction method (sincFold, solid black line) are added as baselines.
    Next to RNA-LLM performances, sequence length distribution for each dataset is shown in blue. 
    \textbf{a}, ArchiveII 5-fold random cross-validation.
    A Wilcoxon test for paired samples with Bonferroni correction indicates that all differences are statistically significant ($P<0.0001$, $N=3,864$). 
    \textbf{b}, bpRNA train-test partitions with controlled homology.
    All differences are statistically significant ($P<0.0001$, $N=1,305$) except for one-hot and RNABERT.\@
    \textbf{c}, bpRNA-new dataset, for RNA families not seen during training.
    All differences are statistically significant ($P<0.0001$, except for RNAErnie versus RNA-MSM and one-hot with $P<0.05$, $N=5,401$). 
    \textbf{d}, PDB-RNA dataset, with RNA sequences extracted from PDB.\@
    Most differences are not statistically significant among RNA-LLM, except for RiNALMo and ERNIE-RNA with respect to RNA-FM, RNABERT and RNAErnie.
    Details of statistical analysis in Supplementary Fig. S1a-d.
    \textbf{e}, Average $F_1$ per motif type on bpRNA-new.
    \textbf{f}, Performance accounting only non-canonical interactions on PDB-RNA.}%
    \label{fig:benchmarks}
    \hrulefill%
\end{figure*}

Figure~\ref{fig:benchmarks} shows the comparative results among the RNA-LLM here reviewed in terms of $F_1$ violin plots. In each plot the results are presented in order, from best (left) to worst (right) median $F_1$.
The best thermodynamic and DL-based predictors were used as baselines in the plots (details in Supplementary Table 1).

Figure~\ref{fig:benchmarks}a shows the results for 5-fold random partitions on the ArchiveII dataset.
It can be seen that most predictors achieve a median performance above $F_1=0.60$, except only for the one based on one-hot encoding ($F_1=0.57$). 
The top-3 RNA-LLM achieve a very high performance ($F_1>0.90$). Those are ERNIE-RNA and RiNALMo, both with $F_1=0.95$, and RNA-FM with $F_1=0.91$. RNAErnie and RNA-MSM achieve an intermediate performance, $F_1=0.76$ and $F_1=0.74$, respectively. Finally, RNABERT is the RNA-LLM with the lowest average performance ($F_1=0.62$). For this random $k$-folding, all RNA-LLM achieve higher performance than the classical prediction method ($F_1=0.61$). Remarkably here, the DL-based method achieves the highest median performance ($F_1=0.97$).
This result shows that the neural architecture used in the classifier has the capability of obtaining high performance in predictions. However, it is well-known that the sequence homology between train and test at random partitions favors methods based on deep learning; thus performance tends to be overly optimistic.

For a deeper analysis, we have extended the experiments on standard datasets having controlled levels of homology between training and testing partitions. Figure~\ref{fig:benchmarks}b shows the test $F_1$  for bpRNA dataset, with a partition at 80\% sequence-identity cutoff between training (TR0+VL0) and testing (TS0).

The results show that now, in a harder testing setup, most methods lowered performance in comparison to random partitions. Here, again, ERNIE-RNA and RiNALMo have both the best median scores in the test set, but in this case with a large drop in performance: $F_1=0.68$ and $F_1=0.64$, respectively. These two RNA-LLM are the only ones that achieve better performance than the classical prediction method 
($F_1=0.53$) and very close performance to the DL-based method ($F_1=0.68$).
With median performance below it, RNA-FM has $F_1=0.52$, RNA-MSM has $F_1=0.42$, and RNAErnie has $F_1=0.41$. Both one-hot and RNABERT reached similar and very poor results, $F_1=0.35$ and $F_1=0.33$, respectively. 

Figure~\ref{fig:benchmarks}c shows the results of testing already trained models with the bpRNA-new dataset, which was designed to evaluate the prediction on new RNA families, never seen during training. 
The performance of all methods has been noticeably reduced, 
being the medians of almost all RNA-LLM below the classical folding method ($F_1=0.67$). The only RNA-LLM with the same performance as the classical one is ERNIE-RNA. In all the other cases, the median $F_1$ is below $0.45$. RiNALMo achieved the same median performance as the DL-based baseline ($F_1=0.44$).
Additionally, note that the one-hot encoding moved up one more position in the ranking, slightly outperforming RNAErnie in this test set. This large dataset was used to analyze the prediction scores per structural motif, as seen in Figure~\ref{fig:benchmarks}e. It can be seen that most models behave similarly for each motif type, with ERNIE-RNA showing a higher performance in stem base-pair predictions. Overall, ERNIE-RNA, RiNALMo and RNA-MSM achieved the best scores for each motif.  

Results on a more challenging dataset, closer to a real-world application, are shown in Figure~\ref{fig:benchmarks}d, where every RNA-LLM was tested in a set of 62 RNA sequences whose secondary structures were inferred from high-resolution RNA X-ray 3D structures from the PDB dataset. 
Results clearly show a hard fall in results for all the RNA-LLM, with median performances below $F_1=0.40$, lower than the classical folding method 
($F_1=0.67$) and the DL-based method ($F_1=0.58$).
Notably in this dataset, the one-hot encoding is within the top-3 methods, outperforming RNA-FM, RNAErnie and RNABERT;\@ with almost the same performance that RNA-MSM, ERNIE-RNA and RiNALMo. It can be said that, for this prediction challenge, no RNA-LLM provides significant improvements of transfer learning over a one-hot encoding. This dataset also contains non-canonical interactions, although most of the methods fail to reach a median $F_1$ score higher than $0.2$ (Fig.~\ref{fig:benchmarks}f), with a slightly better performance for ERNIE-RNA and RiNALMo.

Additionally to the $F_1$ score, we have also calculated for all the datasets analyzed the WL metric, which was proposed to capture structural information between predictions and the corresponding references. Supplementary Table 1 shows that the performances have the same trend to those results reported and analyzed with $F_1$.

\subsection*{Cross-family benchmarks}
\begin{figure*}[t!]
    \centering
    \includegraphics[width=176mm]{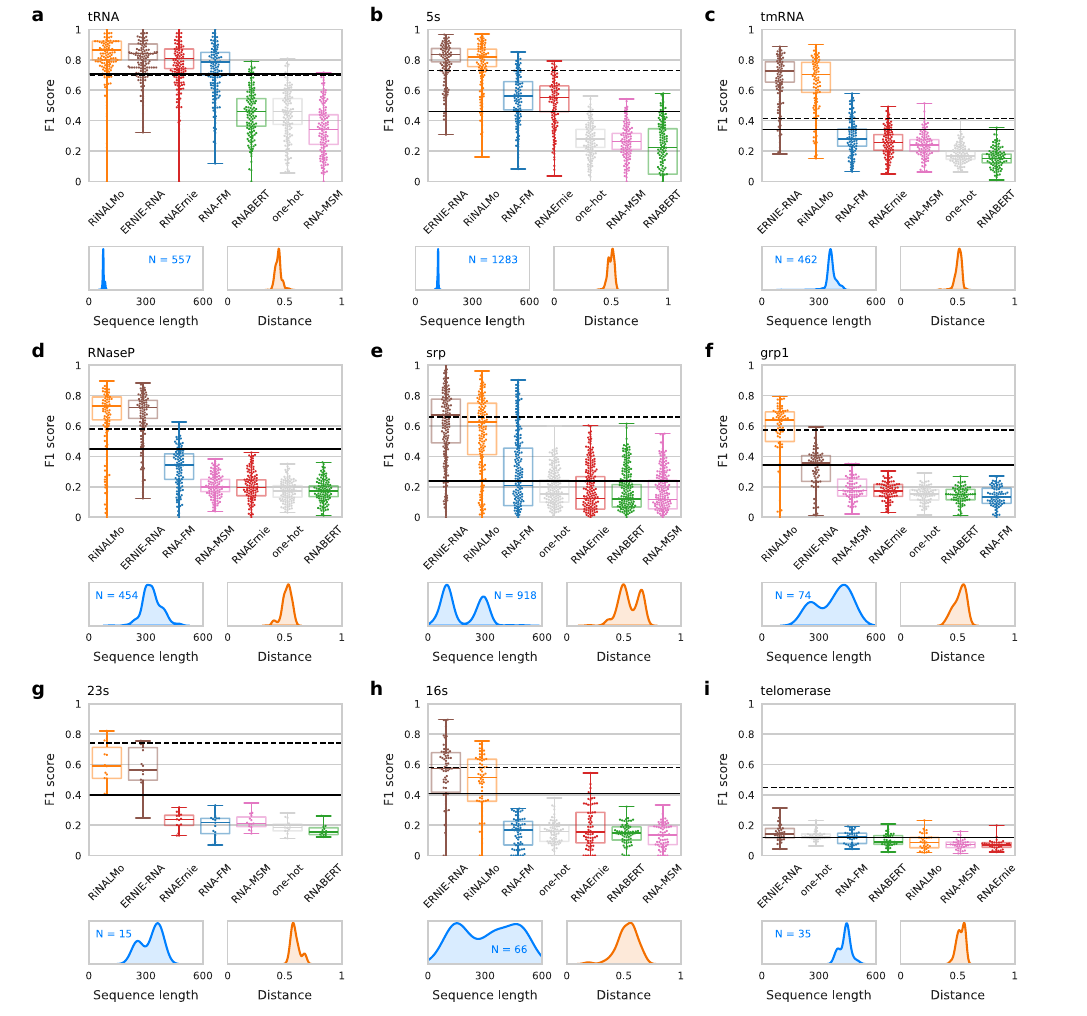}
    \caption{Inter-family structure prediction based on RNA LLM. Performance evaluated on the 9 RNA families of the ArchiveII dataset.
    Each boxplot represents the $F_1$ performance of all methods for a given family in the test set. 
    The thermodynamic prediction method (LinearPartition-V, dashed black) and the DL-based method (sincFold, solid black line) are added as baselines in all plots.
    Below RNA-LLM performances, sequence length distribution for each dataset is shown in blue and distribution of minimum test-train structural distance~\cite{Reuter2010} is shown in orange.
    \textbf{a}, tRNA family.
    A Wilcoxon test for paired samples with Bonferroni correction indicates that all differences are statistically significant ($P<0.0001$, $N=557$), except in the case of RiNALMo with ERNIE-RNA, and RNABERT with one-hot.
    \textbf{b}, 5s family.
    All differences are statistically significant ($P<0.0001$, $N=1283$).
    \textbf{c}, tmRNA family.
    All differences are statistically significant ($P<0.0001$, $N=462$).
    \textbf{d}, RNaseP family.
    Only ERNIE-RNA with RiNALMo, RNA-MSM with RNAErnie, and one-hot with RNABERT are not statistically significant.
    \textbf{e}, srp family.
    Differences for the top 4 methods are statistically significant ($P<0.0001$, $N=918$).
    \textbf{f}, grp1 family.
    Differences for the top 2 methods are statistically significant ($P<0.0001$, $N=74$).
    \textbf{g}, 23s family.
    Top 2 methods are significantly better than the rest of all the methods ($P<0.01$, $N=15$).
    \textbf{h}, 16s family.
    RiNALMo and ERNIE-RNA are statistically different and are also statistically different from the rest of the methods ($P<0.0001$, $N=66$).
    \textbf{i}, telomerase family. ERNIE-RNA and one-hot are statistically different from RNABERT, RiNALMo, RNA-MSM and RNAErnie. Details of statistical analysis in Supplementary Fig. S2.
    }%
\label{fig:familyfold}
\hrulefill%
\end{figure*}

For assessing inter-family performance, a family-fold cross validation in the ArchiveII dataset was performed. That is, one family was left out for testing per cross-validation fold, and the rest of the families were used for training. This eliminates most of the homology to the training set, providing a hard measure of performance and, thus, allowing estimating future performance on novel RNAs that do not belong to any known family.
These results are shown in Figure~\ref{fig:familyfold}, along with 
the distributions of sequence lengths and minimum structural distances~\cite{Reuter2010} between each test sequence and all the sequences in its corresponding training fold.
The best thermodynamic and DL-based predictors are shown as reference for comparison in the plots (detailed results in Supplementary Table 2).

In the case of the tRNA family (Fig.~\ref{fig:familyfold}a), the one with the shortest sequence length as shown in the panel below, most methods achieve performance above $F_1=0.40$. 
RNABERT and one-hot achieved both the same performance here, close to $F_1=0.46$. The four best methods for this family are RiNALMo ($F_1=0.86$), ERNIE-RNA ($F_1=0.84$), RNAErnie ($F_1=0.81$) and RNA-FM ($F_1=0.79$). In this case, which can be considered the easiest one for a cross-family validation due to the small length of sequences and test/train structural distance below 0.5, four of the RNA-LLM 
outperform both the classical method, which achieved $F_1=0.70$, and the DL-based model, which achieved $F_1=0.71$.
For the 5s family (Fig.~\ref{fig:familyfold}b), the results for all RNA-LLM are very similar to the previous family, again due to the length of the sequences (the second mean shortest family) and low structural distance between this family and all the remaining families in the training set. 
In this case, one-hot representation, RNABERT and RNA-MSM are below $F_1=0.26$, while RNAErnie and RNA-FM achieve performances around $F_1=0.55$, and again ERNIE-RNA and RiNALMo achieved both the best results, with $F_1=0.84$ and $F_1=0.82$, respectively, above the classical method 
($F_1=0.73$). The DL-based model achieved $F_1=0.46$ here, below RNAErnie and RNA-FM.
For the tmRNA (Fig.~\ref{fig:familyfold}c) and RNaseP (Fig.~\ref{fig:familyfold}d) families, the conclusions are very similar as before. 
Although ERNIE-RNA and RiNALMo have performance 10\% lower than with the previous families, in these cases achieved even higher performance than 
both the classical and the DL-based method.
In the case of the tmRNA family (Fig.~\ref{fig:familyfold}c), the classical method has 
$F_1=0.42$ and the DL-based has  $F_1=0.34$,
while ERNIE-RNA and RiNALMo have almost twice the performance with $F_1=0.73$. In the case of the RNaseP (Fig.~\ref{fig:familyfold}d) family, those two RNA-LLM methods are more than 10\% better than the classical 
and the DL-based method.
Here, it is important to note that these two families are very similar to each other in average sequence length, around 300 nt, but 3 times longer than tRNA and 5s families.

In the case of the srp family (Fig.~\ref{fig:familyfold}e), there is an important change in the length distribution, which becomes bimodal, and in the structural distances distribution, which clearly exceeds the 0.5 mark. Here, the classical method, ERNIE-RNA and RiNALMo achieve extremely close results %
while all other RNA-LLM %
and the DL-based method  
are far below them. 
ERNIE-RNA has median $F_1=0.67$ while the classical method achieved 
$F_1=0.66$.
Notably, in the case of the grp1 family (Fig.~\ref{fig:familyfold}f), which has larger average length size than srp, RiNALMo is the only LLM that 
clearly outperforms all methods, even the classical one ($F_1=0.64$ versus $F_1=0.57$). In the second position, ERNIE-RNA achieves the same performance as the DL-based method.
In the case of the 23s and 16s families (Figs.~\ref{fig:familyfold}g and~\ref{fig:familyfold}h, respectively), it can be seen that the one-hot representation, RNABERT, RNA-FM, RNA-MSM and RNAErnie embeddings are all below $F_1=0.25$, a very low performance. However, ERNIE-RNA and RiNALMo achieve almost twice that performance with median values near to  $F_1=0.60$, and the classical method achieves 
better results ($F_1=0.74$ for 23s and $F_1=0.58$ for 16s). The DL-based method presents intermediate results: $F_1=0.40$ for 23s and $F_1=0.41$ for 16s.

Notably, this last family has a wide variety of sequence lengths and structural distances.
Finally when the telomerase family (Fig.~\ref{fig:familyfold}i) is used in the test set, all RNA-LLM 
and the DL-based method
achieve extremely poor results below $F_1=0.15$. This family has the longest sequences. Here the classical method has a large advantage over all RNA-LLM 
($F_1=0.45$),
although it is low compared to the other families.

For these dataset partitions we have also calculated the WL metric, showing performance trends similar to those analyzed for $F_1$ score (see details in Supplementary Table 2).

\section*{Discussion} 
We benchmarked 6 RNA-LLM for their ability to predict RNA secondary structure based only on the sequence, in 4 datasets of increasing complexity. First we performed a visual comparison of the LLM embeddings with a UMAP projection, analyzing the distribution and separation of the RNA families. We found that RiNALMo and ERNIE-RNA were the models that could better represent and separate the RNA families in the projection without almost overlap. 
Notably, as Table 1 shows, those RNA-LLM are the biggest ones, pre-trained with the largest and most varied datasets.

We tested each RNA-LLM implementation with exactly the same experimental setup and deep neural network architecture for RNA secondary structure prediction in several benchmark datasets, increasing the difficulty from simple random partitions to very low homology between training and testing partitions. Overall, RiNALMo and ERNIE-RNA maintained an acceptable level of generalization capability even in the hardest test, providing equivalent performance to the classical folding method. Remarkably, in that hardest scenario the trivial one-hot representation achieved a performance similar to the best performing RNA-LLM\@. 
Thanks to the transfer learning paradigm, one would have expected that LLM would help to generalize even with a small training set and prediction model, because all the relevant information was learnt in the embeddings.
That was not the case and it was probably an indication that there is still important information that cannot be fully captured even by the LLM pre-training. The results achieved might be explained by the fact that the structures of sequences in the testing sets were really very different from those seen during pre-training, or they were a small minority, thus the generalization capability of the prediction model could not be really benefited from using a LLM representation.

Precisely regarding the information that may have been incorporated in the pre-training of the RNA-LLMs, it is interesting to compare them with methods that use homology information for predicting RNA secondary structures~\cite{Bernhart2008,Hamada2010,Will2007,Sato2012, Hamada2009, vL2023}. In the case of the dataset closest to a real-world application, PDB-RNA (Figure 2d),  Singh et al.~\cite{Singh2021} reviewed the most relevant methods that use homology information with the following performance: CentroidAlifold~\cite{Hamada2010} $F_1=0.69$, Turbofold II~\cite{Tan2017} $F_1=0.64$, RNAalifold~\cite{Bernhart2008} $F_1=0.66$ and CentroidFold~\cite{Sato2009} $F_1=0.60$. These performances are extremely close to the baseline performance of the pure thermodynamic method ($F_1=0.67$, Figure 2d), and those are actually the highest performance values for this dataset. Instead, here all RNA-LLM show very poor results, with median performances below $F_1=0.40$. Thus, there is still a large performance gap between RNA LLM based and classical methods, independently of explicit homology incorporation. Remarkably, the best DL-based method, without using any homology nor alignment data, has achieved an intermediate performance within the gap mentioned ($F_1=0.58$), even very close to some of the homology-based methods. It is also interesting to compare RNA-LLM trained only with RNA sequences with approaches that incorporate other information in pre-training. Multimodal LLM are interesting since they propose new ways of gaining insights in related data distributions. For example, Evo~\cite{evo} was trained with DNA, considering coding and non-coding sequences. However, since this is fundamentally a generative model trained for sequence design with large contexts, the internal representations obtained for RNA sequences from our datasets were not successful to predict secondary structures. Other methods use abundant DNA sequences in a BERT-like architecture such as DNABERT2~\cite{zhou2023dnabert2}. In this case, the performance of the DNABERT2 model in the datasets here evaluated was just slightly better than RNABERT.\@ From these results, it can be concluded that LLM are still unable to correctly represent the additional information incorporated in the pre-training stage, or if they were able to capture it, it is not trivial to extract this extra information with the models in the downstream tasks.

For the most difficult secondary structure prediction task, we removed from training all the sequences of one specific testing family. In this setup, most of the RNA-LLM achieved a high performance for the two shortest families. However, as the average length of the sequence to be predicted increased, the performance of most RNA-LLM lowered below $F_1=0.50$. Overall, in 4 out of 9 cases RiNALMo and ERNIE-RNA outperformed the classical method. In two cases no differences were observed, and in the last three families the classical method was the best. In the case of the telomerase RNA family, the most different from the other ones, with very few samples in all datasets and the largest average sequence length, the results were all poor, up to three times below the classical method. In any case, in this RNA family the classical method also obtained a performance without practical usefulness, barely reaching $F_1=0.50$. Remarkably, in this hard prediction task, RiNALMo and ERNIE-RNA always outperformed the DL-based prediction method.   

Our study showed that RNA-LLM in combination with a deep neural network for prediction were surprisingly capable of outperforming a classical method in half of the RNA families, even though cross-family prediction was a task historically dominated by thermodynamic methods. Although in this study we have not delved into the analysis of classical methods, because those are outside the scope of the RNA-LLM comparison, it is important to note that their historical superiority is due to a limited understanding of the concept of data-driven learning. This is intuitively associated only to machine learning methods, but all thermodynamics-based methods also have data-driven learning. For machine learning-based methods, humans build a software with rules to be trained automatically from domain examples, thus excluding some of those examples for testing is very simple. Conversely, for classical methods humans learn themselves from data samples, and then build a software with those learned rules. That is: they observe many sequences and structures for years, test and improve their models based on those observations, and even incorporate measurements of real structures as parameters of their models~\cite{Mathews2004}. In these cases, it is much more difficult to simply exclude training samples from testing sets. This is why a new methodology is required to make a fairer comparison with thermodynamic methods.

We found that those RNA-LLM trained in the self-supervised stage with the largest and most varied number of sequences, and with the largest number of trainable parameters (ERNIE-RNA and RiNALMo), were also those that better represented the different RNA families in the projected UMAP space and those that accordingly achieved higher performance overall, very consistently and in all cases.
At the same time, our experimental design clearly revealed the most important challenges that remain unsolved in RNA secondary structure prediction yet. We generated the first systematic benchmarking for this prediction task with LLM.\@ 
Our experimental methodology, the curated datasets and the full source code are a key tool for analysts to navigate the space of available RNA-LLM methods, and constitutes a reference base for developers towards building more efficient prediction methods. We are confident that the criteria and analysis processes defined here can become a benchmark for future systematic investigations of RNA-LLM performance.

\bibliographystyle{plain}
\bibliography{rnallm}

\section*{Funding}
This work was supported by Argentine National Scientific and Technical Research Council (CONICET), ANPCyT (PICT 2022--0086) and UNL (CAI+D 2020 115). We also acknowledge the support of NVIDIA Corporation for the donation of GPU used for this research.

\section*{Data availability}
Curated benchmark datasets of increasing complexity are available in the repository: \href{https://github.com/sinc-lab/rna-llm-folding/tree/main/data}{https://github.com/sinc-lab/rna-llm-folding/tree/main/data}. 

\noindent Moreover, all the embeddings generated in this study are available via Zenodo at: \href{https://doi.org/10.5281/zenodo.13821093}{https://doi.org/10.5281/zenodo.13821093}.

\noindent The source code to reproduce all the experiments and results can be found in: \href{https://github.com/sinc-lab/rna-llm-folding/}{https://github.com/sinc-lab/rna-llm-folding/}.

\end{document}


\eject\pdfpageheight=32cm

\def\maketitle{
    \hrule height 0.5pt
    \vspace{10pt}
    {\noindent\LARGE\bfseries\textsf{\@title} \par}
    \vspace{10pt}
    {\noindent\large\textsf{\@author} \par}
    \vspace{5pt}
    \hrule height 0.2pt
}
\maketitle

\vspace{25pt}

\begin{figure*}[h]
    \centering
    \includegraphics[width=176mm]{fig_S1_v1.pdf}
    \caption{\footnotesize Statistical analysis for RNA-LLM on the RNA secondary structure prediction task for different benchmark datasets of increasing complexity.
    Left: Friedman test and critical difference with post-hoc Nemenyi test with Bonferroni correction (1). 
    Right: Friedman test with Wilcoxon signed-rank test for paired samples with Bonferroni correction as post-hoc test (2). 
    \textbf{a}, ArchiveII 5-fold random cross-validation. 
    \textbf{b}, bpRNA train-test partitions with controlled homology.
    \textbf{c}, bpRNA-new dataset, for RNA families not seen during training. 
    \textbf{d}, PDB-RNA dataset, with RNA sequences extracted from PDB\@.}%
    \label{fig:benchmarks_supp}
\end{figure*}

\newpage
\eject\pdfpageheight=54.5cm

\begin{figure*}[h!]
    \centering
    \includegraphics[width=184mm]{fig_S2_v1.pdf}
    \caption{\footnotesize Statistical analysis for the inter-family structure prediction based on RNA LLM. 
    Left: Friedman test and critical difference with post-hoc Nemenyi test with Bonferroni correction (1). 
    Right: Friedman test with Wilcoxon signed-rank test for paired samples with Bonferroni correction as post-hoc test (2). 
    \textbf{a}, tRNA family. 
    \textbf{b}, 5s family. 
    \textbf{c}, tmRNA family. 
    \textbf{d}, RNaseP family. 
    \textbf{e}, srp family. 
    \textbf{f}, grp1 family. 
    \textbf{g}, 23s family.
    \textbf{h}, 16s family.
    \textbf{i}, telomerase family.}%
    \label{fig:familyfold_supp}
\end{figure*}

\newpage
\eject\pdfpageheight=250mm
\eject\pdfpagewidth=297mm

\begin{table}[!htp]
  \begin{minipage}{15.5cm}
      \textbf{Supplementary Table 1.} Performance metrics for RNA secondary structure prediction models on different benchmark datasets.
      \vspace{0.5mm}
    \end{minipage}
  \small
  \begin{tabular}{lrrrrrrrrr|rrr}\toprule
  \textbf{} &\textbf{} &\multicolumn{2}{c}{\textbf{ArchiveII}} &\multicolumn{2}{c}{\textbf{bpRNA}} &\multicolumn{2}{c}{\textbf{bpRNA-new}} &\multicolumn{2}{c}{\textbf{PDB-RNA}} &\multicolumn{2}{c}{\textbf{Average}} \\\cmidrule{3-12}
  \textbf{} &\textbf{} &$F_1$ &WL &$F_1$ &WL &$F_1$ &WL &$F_1$ &WL &$F_1$ &WL \\\midrule
  \multirow{3}{*}{Classical} &RNAfold &0.59 &0.70 &0.53 &0.68 &0.67 &0.76 &0.68 &0.69 &0.62 &0.71 \\
  &LinearPartition-V &0.61 &0.71 &0.53 &0.68 &0.67 &0.75 &0.67 &0.69 &0.62 &0.71 \\
  &RNAstructure &0.57 &0.68 &0.52 &0.67 &0.64 &0.74 &0.66 &0.67 &0.60 &0.69 \\
  \midrule
  \multirow{2}{*}{Hybrid} &LinearPartition-C &0.67 &0.76 &0.60 &0.75 &0.71 &0.81 &0.68 &0.68 &0.66 &0.75 \\
  &MXfold2 &0.80 &0.85 &0.54 &0.74 &0.61 &0.76 &0.31 &0.47 &0.57 &0.71 \\
  \midrule
  \multirow{3}{*}{Deep learning} &REDfold &0.98 &0.98 &0.69 &0.82 &0.50 &0.73 &0.38 &0.53 &0.64 &0.76 \\
  &UFold &0.93 &0.94 &0.63 &0.74 &0.54 &0.69 &0.58 &0.61 &0.67 &0.75 \\
  &sincFold &0.97 &0.98 &0.68 &0.83 &0.44 &0.72 &0.57 &0.62 &0.67 &0.78 \\
  \midrule
  \multirow{7}{*}{RNA-LLM} &ERNIE-RNA &0.95 &0.95 &0.68 &0.76 &0.67 &0.73 &0.36 &0.40 &0.67 &0.71 \\
  &RNAErnie &0.76 &0.74 &0.41 &0.59 &0.38 &0.55 &0.18 &0.31 &0.43 &0.55 \\
  &RiNALMo &0.95 &0.95 &0.64 &0.76 &0.43 &0.63 &0.37 &0.42 &0.60 &0.69 \\
  &one-hot &0.57 &0.60 &0.35 &0.58 &0.39 &0.58 &0.34 &0.36 &0.41 &0.53 \\
  &RNA-MSM &0.74 &0.71 &0.42 &0.63 &0.39 &0.61 &0.34 &0.33 &0.47 &0.57 \\
  &RNABERT &0.62 &0.64 &0.33 &0.55 &0.36 &0.55 &0.24 &0.33 &0.39 &0.52 \\
  &RNA-FM &0.91 &0.90 &0.52 &0.65 &0.41 &0.58 &0.25 &0.33 &0.52 &0.61 \\
  \bottomrule
  \end{tabular}
  \end{table}

\begin{table}[!htp]
      \label{tab:S2}
      \textbf{Supplementary Table 2.} Performance metrics for RNA secondary structure prediction models on different RNA families.
    \small
    \begin{tabular}{lrrrrrrrrrrrrrrrrrrr|rrr}\toprule
    & &\multicolumn{2}{c}{\textbf{tRNA}} &\multicolumn{2}{c}{\textbf{5s}} &\multicolumn{2}{c}{\textbf{tmRNA}} &\multicolumn{2}{c}{\textbf{RNaseP}} &\multicolumn{2}{c}{\textbf{srp}} &\multicolumn{2}{c}{\textbf{grp1}} &\multicolumn{2}{c}{\textbf{23s}} &\multicolumn{2}{c}{\textbf{16s}} &\multicolumn{2}{c}{\textbf{telomerase}} &\multicolumn{2}{c}{\textbf{Average}} \\\cmidrule{3-22}
    & &$F_1$ &WL &$F_1$ &WL &$F_1$ &WL &$F_1$ &WL &$F_1$ &WL &$F_1$ &WL &$F_1$ &WL &$F_1$ &WL &$F_1$ &WL &$F_1$ &WL \\\midrule
    \multirow{3}{*}{Classical} &RNAfold &0.71 &0.78 &0.69 &0.76 &0.43 &0.60 &0.54 &0.68 &0.67 &0.74 &0.56 &0.68 &0.74 &0.80 &0.53 &0.68 &0.48 &0.62 &0.53 &0.63 \\
    &LinearPartition-V &0.70 &0.76 &0.73 &0.78 &0.42 &0.58 &0.58 &0.70 &0.66 &0.74 &0.57 &0.68 &0.74 &0.82 &0.58 &0.71 &0.45 &0.60 &0.54 &0.64 \\
    &RNAstructure &0.73 &0.79 &0.62 &0.72 &0.40 &0.58 &0.54 &0.68 &0.64 &0.72 &0.52 &0.65 &0.68 &0.78 &0.57 &0.70 &0.46 &0.60 &0.52 &0.62 \\
    \midrule
    \multirow{2}{*}{Hybrid} &LinearPartition-C &0.76 &0.82 &0.78 &0.83 &0.38 &0.60 &0.58 &0.71 &0.68 &0.75 &0.61 &0.73 &0.69 &0.77 &0.66 &0.77 &0.51 &0.65 &0.56 &0.66 \\
    &MXfold2 &0.53 &0.67 &0.58 &0.70 &0.41 &0.59 &0.51 &0.66 &0.61 &0.71 &0.48 &0.62 &0.58 &0.71 &0.51 &0.68 &0.38 &0.57 &0.46 &0.59 \\
    \midrule
    \multirow{3}{*}{Deep learning} &REDfold &0.40 &0.65 &0.49 &0.67 &0.26 &0.56 &0.37 &0.61 &0.17 &0.51 &0.31 &0.60 &0.40 &0.64 &0.36 &0.62 &0.09 &0.53 &0.29 &0.54 \\
    &UFold &0.49 &0.68 &0.38 &0.63 &0.35 &0.58 &0.43 &0.64 &0.20 &0.51 &0.44 &0.63 &0.39 &0.61 &0.29 &0.61 &0.19 &0.54 &0.31 &0.54 \\
    &sincFold &0.71 &0.79 &0.46 &0.65 &0.34 &0.58 &0.45 &0.63 &0.24 &0.53 &0.35 &0.60 &0.40 &0.63 &0.41 &0.63 &0.12 &0.57 &0.35 &0.56 \\
    \midrule
    \multirow{7}{*}{RNA-LLM} &ERNIE-RNA &0.84 &0.87 &0.84 &0.85 &0.73 &0.76 &0.72 &0.76 &0.67 &0.73 &0.36 &0.56 &0.56 &0.69 &0.57 &0.72 &0.14 &0.43 &0.54 &0.64 \\
    &RNAErnie &0.81 &0.84 &0.55 &0.64 &0.26 &0.46 &0.19 &0.48 &0.12 &0.43 &0.17 &0.40 &0.24 &0.52 &0.15 &0.52 &0.07 &0.40 &0.26 &0.47 \\
    &RiNALMo &0.87 &0.88 &0.82 &0.84 &0.70 &0.77 &0.73 &0.74 &0.63 &0.70 &0.64 &0.71 &0.59 &0.67 &0.51 &0.66 &0.09 &0.44 &0.56 &0.64 \\
    &one-hot &0.46 &0.60 &0.28 &0.50 &0.17 &0.35 &0.17 &0.39 &0.15 &0.39 &0.15 &0.37 &0.18 &0.43 &0.16 &0.40 &0.13 &0.36 &0.19 &0.38 \\
    &RNA-MSM &0.34 &0.55 &0.26 &0.48 &0.24 &0.45 &0.20 &0.44 &0.12 &0.41 &0.18 &0.43 &0.21 &0.45 &0.14 &0.45 &0.07 &0.41 &0.18 &0.41 \\
    &RNABERT &0.46 &0.60 &0.22 &0.49 &0.15 &0.41 &0.17 &0.41 &0.12 &0.42 &0.15 &0.41 &0.16 &0.42 &0.15 &0.42 &0.09 &0.39 &0.17 &0.40 \\
    &RNA-FM &0.79 &0.83 &0.56 &0.63 &0.28 &0.56 &0.35 &0.52 &0.21 &0.51 &0.13 &0.42 &0.22 &0.56 &0.17 &0.52 &0.13 &0.43 &0.28 &0.50 \\
    \bottomrule
    \end{tabular}
 \end{table}

\section*{References}
\textbf{1}. J Dem{\v{s}}ar, Statistical comparisons of classifiers over multiple data sets.
\newblock {\em\protect Journal of Machine Learning Research}
\textbf{7}, 1--30 (2006).

\vspace*{2mm}
\noindent \textbf{2}. A Benavoli, G Corani, F Mangili, Should we really use post-hoc tests based on
mean-ranks?
\newblock {\em\protect Journal of Machine Learning Research}
\textbf{17}, 1--10 (2016).